\documentclass{article} 
\usepackage{iclr2024_conference,times}

\usepackage{amsmath,amsfonts,bm}

\def\eqref#1{equation~\ref{#1}}

\def\1{\bm{1}}

\DeclareMathAlphabet{\mathsfit}{\encodingdefault}{\sfdefault}{m}{sl}
\SetMathAlphabet{\mathsfit}{bold}{\encodingdefault}{\sfdefault}{bx}{n}

\usepackage{hyperref}
\usepackage{float}
\usepackage{url,subfigure,amsmath,amssymb,epsfig,verbatim,booktabs,graphicx,epstopdf}
\usepackage{graphicx} 
\usepackage{natbib}  
\usepackage{caption} 
\DeclareCaptionStyle{ruled}{labelfont=normalfont,labelsep=colon,strut=off} 
\usepackage{algorithm}
\usepackage{algorithmic}

\usepackage{tablefootnote}
\usepackage[flushleft]{threeparttable}
\usepackage{lipsum}
\usepackage{graphicx}
\usepackage{mathrsfs}
\usepackage{makecell}
\usepackage{etoolbox}
\usepackage{multirow}
\usepackage{stfloats}
\usepackage{amssymb}
\usepackage{amsmath}
\usepackage{mathtools}
\usepackage{array}
\usepackage{booktabs}
\usepackage{enumitem}
\usepackage{wrapfig}
\usepackage{booktabs}
\usepackage{threeparttable}
\usepackage{graphicx}

\usepackage{xcolor}

\title{Enhancing the Cross-Size Generalization for Solving Vehicle Routing Problems via Continual Learning}

\author{
  Jingwen Li$^{1}$, Zhiguang Cao$^{2}$, Yaoxin Wu$^{3,\dagger}$, Tang Liu$^{1}$ \\
  $^{1}$Sichuan Normal University \quad
  $^{2}$Singapore Management University \quad
  $^{3}$Eindhoven University of Technology \\
  \texttt{\{lijingwen,liutang\}@sicnu.edu.cn} \\
  \texttt{zhiguangcao@outlook.com} \\
  \texttt{wyxacc@hotmail.com} \\
  $^{\dagger}$Corresponding author
}

\begin{document}

\maketitle

\begin{abstract}
Deep models for vehicle routing problems are typically trained and evaluated using instances of a single size, which severely limits their ability to generalize across different problem sizes and thus hampers their practical applicability. To address the issue, we propose a continual learning based framework that sequentially trains a deep model with instances of ascending problem sizes. Specifically, on the one hand, we design an inter-task regularization scheme to retain the knowledge acquired from smaller problem sizes in the model training on a larger size. On the other hand, we introduce an intra-task regularization scheme to consolidate the model by imitating the latest desirable behaviors during training on each size. Additionally, we exploit the experience replay to revisit instances of formerly trained sizes for mitigating the catastrophic forgetting. Extensive experimental results show that the proposed approach achieves predominantly superior performance across various problem sizes (either seen or unseen in the training), as compared to state-of-the-art deep models including the ones specialized for the generalizability enhancement. Meanwhile, the ablation studies on the key designs manifest their synergistic effect in the proposed framework.
\end{abstract}

\section{Introduction}
\label{sec:intro}

Combinatorial optimization problems hold significant practical value to various application domains~\citep{korte2011combinatorial}. Among these, vehicle routing problems (VRPs), exemplified by the traveling salesman problem (TSP) and the capacitated vehicle routing problem (CVRP), stand as quintessential representatives. VRPs aim to find the optimal route for vehicles serving a group of customers in various real-life scenarios, such as parcel pickup/delivery, passenger transportation, and home health care~\citep{baker2003genetic,schneider2014electric}.
Despite the extensive efforts in computer science and operations research, traditional exact and heuristic algorithms still encounter challenges when solving VRPs due to their NP-hard nature~\citep{lenstra1981complexity}. These algorithms often require massive tuning to determine the hand-crafted rules and related hyperparameters. To mitigate this issue, deep (reinforcement) learning based methods have been extensively studied and applied to solve VRPs in recent years~\citep{bengio2021machine,zhang2023review}, which leverage neural networks to automatically learn (heuristic) policies from the experience of solving similar VRP instances. Bolstered by advanced neural networks and training approaches, some of these deep models have achieved competitive or even superior performance to the traditional algorithms~\citep{li2021heterogeneous,li2023learning,kong2024efficient,sun2019using}.

Typically, existing deep models are often trained and evaluated on single-sized problem instances, where they are able to deliver decent and efficient solutions. However, the performance of learned policies diminishes when applied to sizes not encountered during the training phase. This limitation becomes more pronounced as the disparity between the sizes further increases. Such a cross-size generalization issue considerably hinders the applications of deep models, especially given that real-world VRP instances consistently present a diverse range of problem sizes.

To address this issue, we propose a continual learning (CL)~\citep{chen2018lifelong} based framework that sequentially trains a deep model on instances of ascending problem sizes. This approach enables the model to perform favorably across a range of problem sizes, covering both those seen and unseen during the training phase. Specifically, we preserve exemplary models derived from the previous training, and leverage the regularization scheme to retain their knowledge for facilitating the subsequent training. We design two distinct regularization terms in the loss function, i.e., the inter-task and the intra-task regularization terms. During the training on each size, the former aims to transfer the valuable insights from smaller-sized tasks to larger-sized ones, while the latter enables the imitation of the most recent exemplar models. Intuitively, both schemes expedite the training on a newly encountered size, with the aid of previously attained experience in problem solving. Additionally, we tailor an experience replay technique~\citep{rolnick2019experience} to intermittently revisit the instances of previously trained smaller sizes for mitigating the catastrophic forgetting~\citep{french1999catastrophic}. Notably, the proposed continual learning only improves the training algorithm of existing deep models, without altering their original neural architectures. It has a great potential to be deployed with different models, without inducing extra inference time. Experimental results indicate that our approach significantly raises the cross-size generalization performance of deep models for both seen and unseen problem sizes. Furthermore, it generally outperforms the state-of-the-art methods that are specially designed for enhancing the generalizability of deep models, showing the effectiveness of our algorithmic designs.


Accordingly, our contributions are summarized as follows: (1) We propose a model-agnostic continual learning based framework to improve the cross-size generalization capabilities of deep models for VRPs. With a single training session, the proposed approach empowers deep models to deliver promising results for VRPs across a wide range of problem sizes, without incurring extra inference time.
(2) To expedite the training on new sizes, we design the inter-task regularization scheme to facilitate the knowledge transfer from smaller to larger sizes. Alternatively, the intra-task regularization scheme consolidates the model by imitating the most recent exemplar models on the current size. On the other hand, we employ the experience replay to counteract the catastrophic forgetting, retaining the competence of deep model in handling smaller-size instances beyond its training on larger ones. (3) We evaluate our approach on TSP and CVRP across a wide range of sizes (seen or unseen during the training). Results on both synthetic and (real-world) benchmark datasets show that our approach bolsters the cross-size generalization, yielding predominantly superior performance to the state-of-the-art methods specialized for generalizability enhancement.

\section{Related work}
\label{sec:relatedwork}

In this section, we review deep models for VRPs and representative works on enhancing cross-size generalization. Then, we brief on the generic continual learning in the machine learning community.

\paragraph{Deep models for VRPs.} Recent learning based methods, i.e., deep models, have shown promise in solving VRPs by automatically discovering effective policies. 
\citet{vinyals2015pointer} tendered the Pointer network 
to learn constructing TSP solution supervisedly, which was further extended to reinforcement learning~\citep{bello2016neural} and CVRP~\citep{nazari2018reinforcement}. 
Similarly, the graph conventional network (GCN) was leveraged to estimate probabilities of each edge appearing in the optimal TSP solution~\citep{joshi2019efficient}. With recent advances of the self-attention mechanism, the attention model (AM)~\citep{kool2018attention} was tailored from Transformer~\citep{vaswani2017attention} for solving VRPs and recognized as a landmark contribution in this field. The follow-up works diverged by (slightly) restructuring AM or targeting diverse VRP variants~\citep{xin2020multi,li2021deep}. The policy optimization with multiple optima (POMO)~\citep{kwon2020pomo} improved AM by exploiting symmetric rollouts and data augmentation technique, achieving state-of-the-art performance for VRPs. Despite the efficient inference, the above methods usually require heavy post-processing procedures to enhance solution quality, such as sampling~\citep{li2021heterogeneous}, active search~\citep{hottung2021efficient}. Especially, some works attempt to improve the generalization performance of deep models in handling distribution shift~\citep{jiang2022learning,bi2022learning,hottung2021efficient,zhou2023towards}. Instead, this paper aims to enhance the cross-size generalization towards a deep model capable of well solving different-sized VRPs.



\paragraph{Cross-size generalization.} 

The above deep models are often trained to solve single-sized VRP instances for attaining favorable evaluation results on that problem size. However, their performance degenerates when the models are evaluated on sizes unseen during the training. To address this cross-size generalization issue, \citet{lisicki2020evaluating} proposed a curriculum learning method to solve TSP instances spanning a range of problem sizes.
Similarly, \citet{zhang2023neural} utilized the curriculum learning to train a deep model on different-sized TSP,
with the knowledge distillation used for training on the largest TSP. 
Nevertheless, both methods are limited to TSP and lack the versatility in addressing broader VRP variants. 
Instead, \citet{zhou2023towards} worked on improving generalization performance across sizes and distributions, by introducing a meta-learning approach to initialize deep models for rapid adaptation to target VRPs. 
However, its performance is contingent on the heavy base model and tricky meta-learning process, which could suffer from a high training cost in the absence of well pre-trained deep models.


In this paper, we first use continual learning to enhance cross-size. Note that our work is different from the ones attempting to solve large-scaled VRPs, which require extra inefficient training/post-processing for the target size~\citep{qiu2022dimes,sun2023difusco,li2021learning,fu2021generalize,hou2022generalize,zong2022rbg}. Our overarching goal is developing a single model with favorable performance in a broad spectrum of problem sizes, in only a single training session.

\paragraph{Continual learning.}
Continual learning (CL) is advantageous in sequentially learning a stream of relevant tasks by absorbing and accumulating knowledge over them~\citep{hadsell2020embracing}. 
However, CL is generally limited by catastrophic forgetting, where learning a new task usually results in a performance degradation on the old tasks. 
To address this issue, numerous efforts have been devoted in recent years to strike a desirable balance between learning plasticity and memory stability. These works can be broadly categorized into three groups, i.e., regularization-based approaches~\citep{li2017learning} that regularize the current training
with the knowledge acquired in the past training; replay-based approaches~\citep{rebuffi2017icarl} that revisit data distributions of previous tasks; and parameter isolation approaches~\citep{mallya2018packnet} that freeze parameters associated with earlier tasks. Continual learning has widespread applications in visual classification~\citep{he2021online}, semantic segmentation~\citep{michieli2019incremental}, natural language processing~\citep{han2021econet}, to name a few. We direct interested readers to~\citep{de2021continual,parisi2019continual} for more details of CL.
In this paper, we introduce the continual learning into VRP domain,
and empirically testify its potential in training deep models that favorably solve different-sized VRPs. 

\section{Preliminaries and Notations}
\label{sec:formulation}
We first formally describe the vehicle routing problems (VRPs) with the objective of yielding high-quality solutions across a spectrum of problem sizes. 
Then, we present the commonly used encoder-decoder structured deep models for constructing solutions to VRPs in an autoregressive manner.

\subsection{VRP Statement}
Following the literature \citep{kool2018attention,wu2021learning}, we focus on two representative routing problems, i.e., TSP and CVRP, respectively. We define a VRP instance over a graph $G=(V, E)$, where $V$ signifies (customer) nodes and $E$ signifies edges between every two different nodes. With $N$ customer in different locations, TSP aims to find the shortest Hamiltonian cycle of $V=\{v_i\}_1^N$, which satisfies that each node in $V$ is visited exactly once. With an auxiliary depot node $v_0$, CVRP extends TSP by considering a fleet of identical vehicles, each of which traverses locations of customers for serving them. Specifically, each vehicle starts from the depot, serves a subset of customers and ultimately returns to the depot. The  constraint on the route of a vehicle is that the total demand of customers in a route cannot exceed the vehicle capacity and each customer is visited exactly once. 

\noindent\textbf{Objective Function.} The solution (i.e., tour) $\tau^N$ to a VRP instance can be described as a permutation of $N$ nodes in $V$. The objective function is often defined as the tour length. For example, 
the objective function of TSP is $C(\tau^N) = \sum_{\{v_i,v_j\}\in \tau^N} D(v_i, v_j)$, where $D(v_i, v_j)$ means the Euclidean distance between the nodes $v_i$ and $v_j$. In this paper, we focus on optimizing objective values of VRPs across multiple problem sizes. 
By referring various sizes to a series $\{N_1, N_2, ..., N_K\}$, 
the cross-size objective function could be defined as the average value of the expected tour lengths over the $K$ sizes, i.e., $L = \frac{1}{K} \sum_{i=1}^K \mathbb{E}[C(\tau^{N_i})]$, reflecting the overall performance of deep models.



\subsection{Autoregressive deep models for VRPs}

Deep models often learn constructing solutions to TSP instances in an autoregressive manner. Specifically, they model the solution construction procedure of VRPs as a Markov Decision Process~(MDP). Then the encoder-decoder structured policy network is adopted to sequentially construct solutions. More specific, the encoder projects problem-specific features into high-dimensional node embeddings for informative representation learning. Afterwards, the decoder sequentially constructs a solution $\tau^{N_i}$ for a TSP instance of problem size $N_i$, conditioned on the updated node embeddings and partial tour at each step. 
During solution construction, the decoder selects a node  $a_{t_c}$ at step $t_c$, with all constraints satisfied by masking the invalid nodes. A feasible solution is constructed until all customer nodes are selected, which is expressed by the factorization below,
\begin{equation}
    p_{\theta} (\tau^{N_i} | G) = \prod_{t_c=1}^{T_c} p_{\theta} (a_{t_c} | a_{1:{t_c}-1}, G),
    \label{eq:construction}
\end{equation}
where $p_{\theta}$ and $T_c$ signifies the policy network and the total number of decoding steps, respectively. In particular, $T_c\!=\!{N_i}$ for TSP, and $T_c\!\geq\! {N_i}$ for CVRP as the depot node can be visited multiple times.

\section{Methodology}
\label{sec:model}


\begin{figure}
\centering 
\setlength{\belowcaptionskip}{-0.5cm}
	\includegraphics[width=\textwidth]{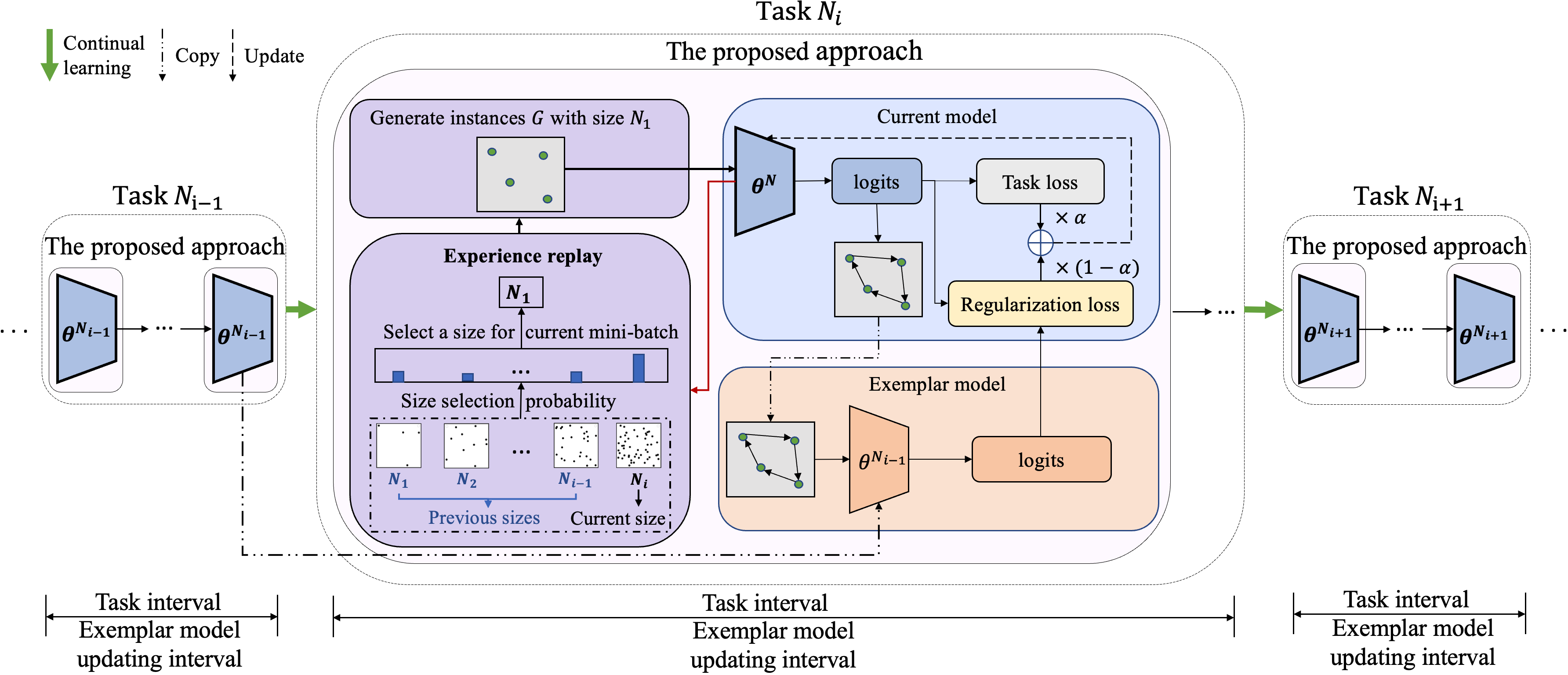}
	\caption{The illustration of the proposed framework with inter-task regularization. {For each mini-batch training during current task interval, we employ 1) experience replay to sample a size from formerly trained sizes and current one, and generate instances with that sampled size; 2) inter-task regularization to foster the current model to emulate an exemplary model for knowledge retention.}}
	\label{fig:framework} 
\end{figure}
Continual learning has emerged as a powerful approach for handling sequential tasks, which enables deep models to progressively retain and accumulate knowledge from evolving data streams. As illustrated in Figure \ref{fig:framework}, we harness CL to enhance the cross-size generalization capability of an autoregressive deep model $\theta$ (e.g., POMO \citep{kwon2020pomo}), by sequentially training it on VRP instances of ascending problem sizes $\{N_1, N_2, ..., N_K\}$. To ensure general favorable performance across the size spectrum, each size (i.e., task) $N_i$ ($i=1, ..., K$) is considered equally important and trained with the same task interval, which is defined as $E_p = E / K$ epochs where $E$ denotes the total training epochs of CL.
In each task interval, the model is trained on each size to optimize the task-specific objective. Meanwhile, our approach exploits experience replay strategy to revisit instances of previously trained smaller sizes, so as to mitigate the catastrophic forgetting. Moreover, the inter-task or intra-task regularization scheme foster the model in current interval to emulate an exemplary model derived from previous or current interval, so as to inherit the previous learned knowledge. In this sense, our CL approach facilitates a coherent continuum of learning across varying problem sizes, which is elaborated in the following sections. 

\subsection{Experience replay}
Experience replay has shown promise to alleviate the catastrophic forgetting issue in continual learning, with the basic logic of reminding the model about the policy learned for previous tasks. A typical experience replay technique is to maintain a small memory buffer of training samples. These samples are collected from the past tasks and replayed during the training on subsequent tasks. 
Given that existing deep models for VRPs are generally trained with random instances~\citep{kool2018attention,kwon2020pomo}, we propose to randomly generate instances of smaller sizes on the fly. Such real-time memory buffer is able to reflect the instance patterns in previous tasks and raise the memory efficiency, when the deep model is trained on a newly encountered larger size. 

During the training on the problem size $N_i$ ($i>1$), we harness a sampling strategy to either randomly select a size from the set of formerly trained sizes $N_{pre}=\{N_1,...,N_{i-1}\}$, or deterministically select the current size $N_i$. This strategy is devised to ensure that the deep model is primarily trained on the current task, i.e., the VRP with a larger size and higher complexity than the previous ones. Meanwhile, it ensures the competence of the deep model is retained for well solving previous tasks, i.e., the VRPs with smaller sizes but subjected to the catastrophic forgetting. 
To this end, we sample problem sizes in mini-batches during the training on size $N_k$, by assigning a higher probability to select $N_i$ and a lower probability to uniformly select one from $N_{pre}$, such that,
\begin{equation}
\begin{split}
& N_k=
\begin{cases}
N_i, & 
\text{if}\ \epsilon < 0.5 \\
N_j \sim U(N_{pre}), & \text{otherwise} 
\end{cases}\\
\end{split}
\label{eq:Er}
\end{equation}
where $\epsilon \in (0,1)$ is a random number. Specially, only the size $N_1$ is involved in the first task.

\begin{figure}
\centering 
\setlength{\belowcaptionskip}{-0.5cm}
	\includegraphics[width=\textwidth]{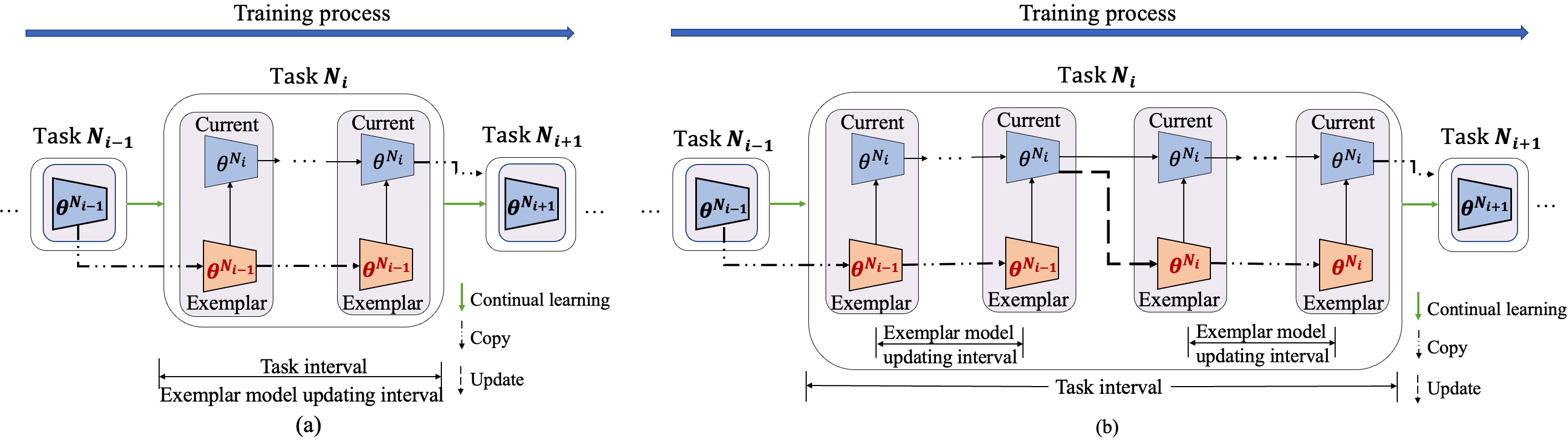}
	\caption{Regularization with two exemplar model updating strategies. {(a) inter-task: exemplar model is updated after training on a whole task; (b) intra-task: exemplar model is updated multiple times during training on a task for concentrating more on newly encountered (larger) size.}}
	\label{fig:KD} 
\end{figure}

\subsection{Regularization schemes}
During the training process, we employ favorable models trained previously as the \emph{exemplar} ones to infuse the \emph{current} model with a wealth of knowledge in VRP solving, with the goal to guide the training on the newly encountered size. Specifically, we design two distinct terms in the loss function, i.e., inter-task regularization term and intra-task regularization term, respectively, with different update rule for the exemplar model.
Note that only one regularization scheme can be used in our CL approach to keep a stable update of the exemplar model throughout the training. 

\noindent{\textbf{Inter-task regularization scheme.}} As shown in Figure \ref{fig:KD}(a), the inter-task regularization scheme aims to retain knowledge derived from the past training on smaller sizes {for achieving generalization across various sizes}. Specifically, when training on size $N_i$, the current model $\theta^{N_i}$ is thoughtfully guided by the exemplar model $\theta^{N_{i-1}}$ meticulously trained on the preceding size $N_{i-1}$. In this fashion, the exemplar model is updated after training on each size, with the update interval equal to the task interval, i.e., $E_{inter} = E_p$. This strategy encourages the current model to imitate the solution construction policy learned by the exemplar model. Given a training instance $G$ with size $N_i$ and a tour $\tau_{\theta^{N_i}}$ constructed by $\theta^{N_i}$, we leverage the exemplar model $\theta^{N_{i-1}}$ to engender the same tour, resulting in the probability distribution $p_{\theta^{N_{i-1}}} (\tau_{\theta^{N_i}} | G)$. 
The inter-task regularization loss $\mathcal{L}_{R_{inter}}$ is defined as the similarity between probability distributions derived by $\theta^{N_i}$ and $\theta^{N_{i-1}}$ over a mini-batch of instances $\{G_b\}_{b=1}^B$, which is calculated by the Kullback-Leibler divergence as below,
\begin{equation}
    \mathcal{L}_{R_{inter}} = \frac{1}{B}\sum_{b=1}^B \sum_{a_j \in \tau_{\theta^{N_i}}^b} p_{\theta^{N_{i-1}}} (a_j | G_b) (log p_{\theta^{N_{i-1}}} (a_j | G_b) - log p_{\theta^{N_i}} (a_j | G_b)).
    \label{eq:inter-task}
\end{equation}
Particularly, for training on the first size, a pre-trained backbone model (such as POMO \citep{kwon2020pomo}) on size $N_1$ could be used to serve as the exemplar model in Eq. (\ref{eq:inter-task}). 

\noindent{\textbf{Intra-task regularization scheme.}} As illustrated in Figure \ref{fig:KD}(b), the intra-task regularization scheme concentrates more on consolidating the recently learned knowledge, thereby updating the exemplar model more frequently than inter-task scheme.
Specifically, during the training on size $N_i$ in the task interval, we update the exemplar model $M$ times with an even update interval $E_{intra} = E_p / M$. Given the current epoch $e$, the training of the model ${\theta^{N_i}_e}$ is guided by the most recent exemplar model $\theta^{N_i}_m$ ($m=1,2,...,M$). Accordingly, the intra-task regularization loss $\mathcal{L}_{R_{intra}}$ over a mini-batch of instances $\{G_b\}_{b=1}^B$ is formulated as follows,
\begin{equation}
\setlength{\abovedisplayskip}{5pt}
\setlength{\belowdisplayskip}{5pt}
    \mathcal{L}_{R_{intra}} = \frac{1}{B}\sum_{b=1}^B \sum_{a_j \in \tau_{\theta^{N_i}_e}^b} p_{\theta^{N_i}_m} (a_j | G_b) (log p_{\theta^{N_i}_m} (a_j | G_b) - log p_{\theta^{N_i}_e} (a_j | G_b)).
    \label{eq:intra-task}
\end{equation}
In contrast to inter-task regularization scheme using exemplar model from previous size, {intra-task scheme adopts the one that has already been exposed to the intricacies of a new size, which could assimilate more generalized and resilient knowledge to boost the training efficiency and is preferred for generalizing to unseen larger sizes.}
However, the deep model cannot be sufficiently trained on a new size in the initial stage of a task interval. Thus we employ the finally well-established model $\theta^{N_{i-1}}_M$ on the last size as the exemplar, during the first $E_{intra}$ epochs in the current task interval. 

Finally, the deep model is trained with the objective of minimizing a weighted combination of the regularization term $\mathcal{L}_{R}$ {(i.e., $\mathcal{L}_{R_{inter}}$ for inter-task regularization and $\mathcal{L}_{R_{intra}}$ for intra-task regularizatio)} and the original task loss $\mathcal{L}_T$, i.e., $\mathcal{L} = \alpha \mathcal{L}_{R} + (1-\alpha) \mathcal{L}_T,$, where $\alpha \in [0,1]$. Taking inter-task regularization term as an example, the task loss is formulated as below,
\begin{equation}
    \mathcal{L}_T = \mathbb{E}_{G \sim N_k, \tau^{N_k} \sim p_{\theta^{N_i}}(\tau^{N_k} | G)} [C(\tau^{N_k} | G)],
    \label{eq:task}
\end{equation}
where the training instances are sampled with the selected size $N_k$ via the experience replay strategy, and the tour $\tau^{N_k}$ is engendered via the current network $\theta^{N_i}$ according to Eq. (\ref{eq:construction}). The task loss is used to update the deep model by REINFORCE~\citep{williams1992simple}, which is a commonly applied reinforcement learning algorithm in VRP literature~\citep{kool2018attention,kwon2020pomo}.



\subsection{Training algorithm}
\label{sec:training_algorithm}
\begin{algorithm}[t]
\caption{Model training by continual learning}
\label{ppo}
\begin{algorithmic}[1]
\REQUIRE An ascending sequence of problem sizes $N_1, N_2, ..., N_K$ with equal space $n$; a pre-trained backbone model (e.g., POMO) parameterized by $\theta^{N_1}$ on size $N_1$; 
\FOR{$\text{epoch}\ e = 1, 2,..., E$}
    \STATE Compute the size $N_i = N_1 + n*(e \ \% \ E_p)$ of current task;
    \FOR{$\text{step}\ t=1,2,...,T$}
    \STATE Pick a size $N_k, k=1, ..., i$ according to Eq. (\ref{eq:Er});
    \STATE Randomly generate a batch of training instances with size $N_k$;
    \STATE Let model $\theta$ (e.g., $\theta^{N_i}$ for inter-task regularization) sample tours $\tau_{\theta}^b$ for each $\{G_b\}_{b=1}^B$;
    \STATE Compute $\nabla \mathcal{L}_R$ using Eq. (\ref{eq:inter-task}) for inter-task regularization or Eq. (\ref{eq:intra-task}) for intra-task one;
    \STATE Compute $\nabla \mathcal{L}_T$ using Eq. (\ref{eq:task});
    \STATE $\theta \gets \theta + \eta \nabla \mathcal{L}$ where $\nabla \mathcal{L} \gets \alpha {\nabla\mathcal{L}_{R}} + (1-\alpha) \nabla\mathcal{L}_T$.
    \ENDFOR
\ENDFOR
\end{algorithmic}
\end{algorithm}

We outline the training procedure of the proposed CL approach in Algorithm 1, where the model is sequentially trained using instances with ascending problem sizes $N_1, ..., N_K$. Particularly, starting with the training on size $N_2$, the experience replay strategy plays the role to retain the competence in tackling smaller-size instances when addressing a new larger one. Moreover, the regularization scheme, i.e., either inter-task or intra-task, is smoothly incorporated during the whole training process, transferring previous valuable knowledge to facilitate the subsequent training. In this sense, the proposed approach is expected to endow the deep models with strong cross-size generalization ability so that they could perform favorably across a wide range of sizes.


\section{Experiments}
\label{sec:exp}

To demonstrate the generality and effectiveness of the proposed framework, we apply it to two well-known and strong deep models, i.e., POMO~\citep{kwon2020pomo} and ELG~\citep{gao2023towards}, referred to as Ours-POMO and Ours-ELG, respectively. We conduct comprehensive experiments on two representative routing problems, i.e., TSP and CVRP~\citep{kool2018attention,wu2021learning}, respectively. 

\paragraph{Training setups.} We adhere to most of the setups in POMO and ELG. For our approach, we set the ascending problem sizes $\{\!N_1,\! N_2,\! ...,\! N_K\!\}$ to $\{60, \!70,\! ..., \!150\}$ with $K\! =\! 10$. Note that these sizes could be flexibly adjusted to other incremental values. Regarding Ours-POMO, we set the training epochs to $E\! =\! 2000$, with instances of each size trained for $E_p\! =\! 200$ epochs, ensuring robust performance across the wide range of problem sizes. The update interval of the exemplar model is set to $E_{inter}\!=\!200$ epochs for the inter-task regularization and $E_{intra}\!=\!25$ epochs for the intra-task regularization. Regarding Ours-ELG, we follow the original design of ELG and set the training epochs to $E\! =\! 500$. Accordingly, the update interval of the exemplar model is set to $E_{\text{inter}} = 50$ epochs for the inter-task regularization and $E_{\text{intra}} = 10$ epochs for the intra-task regularization. Both Ours-POMO and Ours-ELG use a batch size of 64 (32 when the sizes exceed 100) for both TSP and CVRP.

\paragraph{Inference setups.} Complying with the established convention~\citep{kool2018attention}, we randomly generate instances following the uniform distribution for both seen and unseen problem sizes during the training phase. Pertaining to the former, we select the three most representative sizes from the set of $K$ training sizes aforementioned, encompassing the minimum size of 60 (with 10,000 instances), the median size of 100 (with 10,000 instances), and the maximum size of 150 (with 1,000 instances). Pertaining to the latter, we consider three larger unseen sizes, i.e., 200, 300 and 500 (with 128 instances for each), to further assess the generalizability. We conduct all experiments including the training and evaluation on a Linux server equipped with TITAN XP GPUs (with 12 GB memory) and Intel Xeon E5-2660 CPUs at 2.0 GHz. Our dataset and code in Pytorch will be made available. 
\begin{table*}[ht]
\caption{Comparison results on TSP and CVRP (seen scales).}
\centering
\resizebox{0.893\textwidth}{!}{  
\begin{threeparttable}
\begin{tabular}{cl||ccc|ccc|ccc|c}
\toprule
\multicolumn{2}{c||}{\multirow{2}{*}{{{Method}}}} & \multicolumn{3}{c|}{{Test on N=60}} & \multicolumn{3}{c|}{{Test on N=100}} & \multicolumn{3}{c|}{{Test on N=150}} & Average of   \\
\multicolumn{2}{c||}{} & Obj. & Gap & Time & Obj. & Gap & Time & Obj. & Gap & Time & Total costs \\ 
\midrule
\multirow{20}*{\rotatebox{90}{TSP}} 
& Concorde & 6.1729  &  -  & (7m)&  7.7646 &  -  & (1.7h)  & 9.3462 & -  & (22m) & 7.7612\\
& LKH3  & 6.1729  &  0.00\%  & (14m) &  7.7646 &  0.00\%  & (9.8h)  & 9.3462 & 0.00\%  & (2.1h) & 7.7612 \\
\cmidrule(lr){2-12}
&   AMDKD-POMO$^*$  & 6.1828 & 0.16\%  & 36s & 7.7930 & 0.37\%  & 2m & 9.4539 & 1.15\%  & 33s &7.8092 \\
\cmidrule(lr){2-12}
&   POMO-60 & {\textbf{6.1746}} & {\textbf{0.03\%}}  & $\sim$ & 7.8050 & 0.52\%  & $\sim$& 9.5909  & 2.62\%  & $\sim$ & 7.8568  \\
&   POMO-100 & 6.1768 & 0.06\%  & $\sim$ & {\textbf{7.7753}} & {\textbf{0.14\%}}  & $\sim$ & 9.3987 & 0.56\%  & $\sim$ &7.7836  \\
&   POMO-150 & 6.1928 & 0.32\%  & $\sim$ & 7.7875 & 0.30\%  & $\sim$ & {\textbf{9.3812}} & {\textbf{0.36\%}}  & $\sim$ &7.7868  \\
\cmidrule(lr){2-12}
&    {POMO-random} & 6.1778 & 0.08\%  & $\sim$ & 7.7823 & 0.23\%  & $\sim$ & 9.3937 & 0.51\%  & $\sim$ &7.7846  \\
\cmidrule(lr){2-12}
&   AMDKD-POMO  & 6.1820 & 0.15\%  & $\sim$ & 7.7916 & 0.35\%  & $\sim$ & 9.4473 & 1.08\%  & $\sim$ &7.8070 \\
&   Omni-POMO$^\ddagger$ &6.2351 &1.01\% & 34s & 7.8650 & 1.29\%  & 2.5m & 9.4958 & 1.60\%  & 37s &7.8653 \\
\cmidrule(lr){2-12}
&   Ours-POMO-inter & \textbf{\emph{6.1763}} & \textbf{\emph{0.06\%}} & 36s & \textbf{\emph{7.7802}} & \textbf{\emph{0.20\%}}  & 2m & 9.3912 & 0.48\%  & 33s &\textbf{\emph{7.7826} }\\
&   Ours-POMO-intra   & 6.1767 & 0.06\% & $\sim$ & 7.7807 & 0.21\%  & $\sim$ & \textbf{\emph{9.3891}}&  \textbf{\emph{0.46\%}} & $\sim$ &\textbf{7.7822}\\
\cmidrule(lr){2-12}
&   ELG-60 & 6.1772 & 0.07\% & 37s & 7.8098 & 0.58\%  & 1.3m & 9.5593 & 2.28\%  & 15s & 7.8388  \\
&   ELG-100 & 6.1807 & 0.13\%  & $\sim$ & 7.7822 & 0.23\%  & $\sim$ & 9.4131 & 0.72\%  & $\sim$ & 7.7920  \\
&   ELG-150 & 6.1864 & 0.22\%  & $\sim$ & 7.7888 & 0.31\%  & $\sim$ & 9.4036 & 0.61\%  & $\sim$ & 7.7929  \\
\cmidrule(lr){2-12}
&    {ELG-random} & 6.1816 & 0.14\%  & $\sim$ & 7.7876 & 0.30\%  & $\sim$ & 9.4120 & 0.70\%  & $\sim$ & 7.7937  \\
\cmidrule(lr){2-12}
&   Ours-ELG-inter & 6.1789 & 0.10\%  & $\sim$ & 7.7867 & 0.29\%  & $\sim$ & 9.4078 & 0.66\%  & $\sim$ & 7.7911  \\
&   Ours-ELG-intra & 6.1796 & 0.11\%  & $\sim$ & 7.7871 & 0.29\%  & $\sim$ & 9.4053 & 0.63\%  & $\sim$ & 7.7907  \\
\midrule
\midrule
\multirow{20}*{\rotatebox{90}{CVRP}} 
& HGS & 11.9471  &  -  & (15.3h) &  15.5642 &  -  & (25.6h)  & 19.0554 & -  & (6.2h) & 15.5222 \\
& LKH3 & 11.9694  &  0.19\%  & (3.5d) &  {15.6473} &  0.53\%  & (6.5d)  & 19.2208 & 0.87\%  & (13h) &15.6125 \\
\cmidrule(lr){2-12}
&   AMDKD-POMO$^*$  & 12.3561 & 3.42\%  & 56s & 15.8854 & 2.06\%  & 3m & 19.8395 & 4.12\%  & 33s &16.0270 \\
\cmidrule(lr){2-12}
&   POMO-60 & \textbf{12.0656} & \textbf{0.99\%}  & $\sim$ & 16.0914 & 3.39\%  & $\sim$ & 20.2573 & 6.31\%  & $\sim$ & 16.1381  \\
&   POMO-100 & 12.2531 & 2.56\%  & $\sim$ & \textbf{15.7544} &  \textbf{1.22\%}  & $\sim$ &19.6856 & 3.31\%  & $\sim$ & 15.8977 \\
&   POMO-150 & 12.4322 & 4.06\%  & $\sim$ & 15.8924 & 2.11\%  &$\sim$ & \textbf{19.3683} & \textbf{1.64\%}  & $\sim$ &15.8976 \\
\cmidrule(lr){2-12}
&    {POMO-random} & 12.2758 & 2.75\%  & $\sim$ & 15.7942 & 1.48\%  & $\sim$ & 19.6121 & 2.92\%  & $\sim$ &15.8940  \\
\cmidrule(lr){2-12}
&   AMDKD-POMO  & 12.1487 & 1.69\%  & $\sim$ & 15.8119 & 1.72\%  & $\sim$ & 19.5280 & 2.48\%  & $\sim$ &15.8362 \\
&   Omni-POMO$^\ddagger$  & 12.2996 & 2.95\%  & 45s & 15.9878 & 2.72\%  & 2.5m & 19.5975 & 2.85\%  & 45s &15.9616 \\
\cmidrule(lr){2-12}
&   Ours-POMO-inter  & \textbf{\emph{12.0672}} & \textbf{\emph{1.00\%}}  & 56s & 15.7903 & 1.45\%  & 3m & 19.4226 & 1.93\%  & 33s &\textbf{\emph{15.7600}} \\
&   Ours-POMO-intra & 12.0680 & 1.01\%  & $\sim$ & \textbf{\emph{15.7867}}& \textbf{\emph{1.43\%}}  & $\sim$ &\textbf{\emph{19.4040}} & \textbf{\emph{1.83\%}}  & $\sim$ &\textbf{15.7529} \\
\cmidrule(lr){2-12}
&   ELG-60 & 12.0975 & 1.26\%  & 56s & 15.9834 & 2.69\%  & 3.2m & 19.8061 & 3.94\%  & 52s & 15.9557  \\
&   ELG-100 & 12.1397 & 1.61\%  & $\sim$ & 15.8382 & 1.76\%  & $\sim$ & 19.5446 & 2.57\%  & $\sim$ & 15.8408  \\
&   ELG-150 & 12.1920 & 2.05\%  & $\sim$ & 15.8862 & 2.07\%  & $\sim$ & 19.5197  & 2.44\%  & $\sim$ & 15.8660  \\
\cmidrule(lr){2-12}
&    {ELG-random} & 12.1612 & 1.79\%  & $\sim$ & 15.8717 & 1.98\%  & $\sim$ & 19.5405 & 2.55\%  & $\sim$ & 15.8578  \\
\cmidrule(lr){2-12}
&   Ours-ELG-inter & 12.1073 & 1.34\%  & $\sim$ & 15.8692 & 1.96\%  & $\sim$ & 19.5336 & 2.51\%  & $\sim$ & 15.8367 \\
&   Ours-ELG-intra & 12.1110 & 1.37\%  & $\sim$ & 15.8617 & 1.91\%  & $\sim$ & 19.5208 & 2.44\%  & $\sim$ & 15.8312  \\
\bottomrule
\end{tabular}
\begin{tablenotes}
    \item \textbf{Bold} and \textbf{\emph{italics}} refer to the best and the second-best performance, respectively, among all deep models. 
    \item[$\sim$] The inference time of a method is equal to that of the preceding method in the row above, since those deep models except for Omni-POMO utilize the original POMO architecture and result in the same inference efficiency. 
    \item[$\ddagger$] The training size range of Omni-POMO is [50, 200], which is broader than  our [60, 150]. 
\end{tablenotes}
\end{threeparttable}
}
\label{tab:main}
\end{table*}

\begin{table*}[ht]
\caption{Generalization results on TSP and CVRP (unseen scales).}
\centering
\resizebox{0.876\textwidth}{!}{%
\begin{threeparttable}
\begin{tabular}{cl||ccc|ccc|ccc|c}
\toprule
\multicolumn{2}{c||}{\multirow{2}{*}{{{Method}}}} & \multicolumn{3}{c|}{{Test on N=200}} & \multicolumn{3}{c|}{{Test on N=300}} & \multicolumn{3}{c|}{{Test on N=500}} & Average of   \\
\multicolumn{2}{c||}{} & Obj. & Gap & Time & Obj. & Gap & Time& Obj. & Gap & Time & Total costs \\ 
\midrule
\multirow{20}*{\rotatebox{90}{TSP}} 
& Concorde & 10.6683  &  -  & (8m)&  12.9534 &  -  & (11m)  & 16.5219 & -  & (17m) & 13.3812\\
& LKH3  & 10.6683  &  0.00\%  & (25m) &  12.9534 &  0.00\%  & (47m)  & 16.5219 & 0.00\%  & (1.2h) & 13.3812 \\
\cmidrule(lr){2-12}
&   AMDKD-POMO$^*$  & 10.9651 & 2.78\%  & 10s & 13.9793 & 7.92\%  & 33s & 19.4197 & 17.54\%  & 2.5m &14.7880 \\
\cmidrule(lr){2-12}
&   POMO-60 & 11.3360 & 6.27\%  & $\sim$ & 14.8162  & 14.38\%  & $\sim$ & 20.5835  &24.58\%  &$\sim$& 15.5786 \\
&   POMO-100 & 10.8464 & 1.67\%  & $\sim$ & 13.8730 & 7.10\% & $\sim$ & 20.1985 & 22.25\%  &$\sim$ &14.9726  \\
&   POMO-150 &10.7752 & 1.00\%  & $\sim$ & 13.2922 & 2.62\%  & $\sim$ & {18.0793} & {9.43\%}  &$\sim$ &{14.0489}  \\
\cmidrule(lr){2-12}
&    {POMO-random} & 10.8397 & 1.61\%  & $\sim$ & 13.8212 & 6.70\%  & $\sim$ & 19.0881 & 15.53\%  & $\sim$ &14.5830  \\
\cmidrule(lr){2-12}
&   AMDKD-POMO  & 10.9054 & 2.22\%  & $\sim$ & 13.4472 & 3.81\%  & $\sim$ & 18.4477 & 11.66\%  & $\sim$ &14.2668 \\
&   Omni-POMO$^\ddagger$  & 10.8923 & 2.10\%  & 11s & 13.4044 & 3.48\%  & 33s & 17.8146 & 7.82\%  & 2.6m & 14.0371 \\
\cmidrule(lr){2-12}
&   Ours-POMO-inter  &\textbf{\emph{10.7696}} & \textbf{\emph{0.95\%}} & 10s & 13.2723 & 2.46\%  & 33s & 17.8332& 7.94\%  & 2.5m & 13.9584 \\
&   Ours-POMO-intra  & \textbf{10.7531} & \textbf{0.80\%}  & $\sim$ & \textbf{13.2172} & \textbf{2.04\%}  & $\sim$ & 17.7517 & 7.44\%  & $\sim$ & 13.9073 \\
\cmidrule(lr){2-12}
\cmidrule(lr){2-12}
&   ELG-60 & 11.0613 & 3.68\%  & 4s & 13.8150 & 6.65\%  & 7s & 18.2116 & 10.23\%  & 15s & 14.3626  \\
&   ELG-100 & 10.8280 & 1.50\%  & $\sim$ & 13.4278 & 3.67\%  & $\sim$ & 17.7035 & 7.15\%  & $\sim$ & 13.9864  \\
&   ELG-150 & 10.7821 &  1.07\%  & $\sim$ & 13.2802 & 2.52\%  & $\sim$ & 17.4444 & 5.58\%  & $\sim$ & 13.8356  \\
\cmidrule(lr){2-12}
&    {ELG-random} & 10.8131 & 1.36\%  & $\sim$ & 13.3886 & 3.36\%  & $\sim$ & 17.6597 & 6.89\%  & $\sim$ & 13.9538  \\
\cmidrule(lr){2-12}
&   Ours-ELG-inter  & 10.7808 & 1.06\%  & $\sim$ & 13.2643 & 2.40\%  & $\sim$ & \textbf{\emph{17.3802}} & \textbf{\emph{5.20\%}}  & $\sim$ & \textbf{\emph{13.8084}}  \\
&   Ours-ELG-intra & 10.7770 & 1.02\%  & $\sim$ & \textbf{\emph{13.2360}} & \textbf{\emph{2.18\%}}  & $\sim$ &\textbf{17.3045} & \textbf{4.74\%}  & $\sim$ &\textbf{13.7725} \\
\midrule
\midrule
\multirow{20}*{\rotatebox{90}{CVRP}} 
& HGS & 21.9737  &  -  & (1.1h) &  25.8417 &  -  & (1.6h)  & 31.0308 & - & (2.5h) &26.6514 \\
& LKH3 & 22.2146  &  1.10\%  & (2.4h) &  26.2184 &  1.46\%  & (3.2h)  & 31.5213 & 1.58\% & (5.3h) &26.6514 \\
\cmidrule(lr){2-12}
&   AMDKD-POMO$^*$  & 23.8507 & 8.54\%  & 12s & 30.7218 & 17.18\%  & 38s & 48.1260 & 52.68\%  & 3m &34.2328 \\
\cmidrule(lr){2-12}
&   POMO-60 & 24.0638 & 9.51\%  & $\sim$ & 29.6416 & 14.71\%  & $\sim$ & 38.8480 & 25.19\%  & $\sim$ &30.8511  \\
&   POMO-100 & 23.2783 & 5.94\%  & $\sim$ & 28.9372 & 11.98\%  & $\sim$ & 37.9132 & 22.18\%  & $\sim$ &30.0429  \\
&   POMO-150 & \textbf{\emph{22.4706}} & \textbf{\emph{2.26\%}}  & $\sim$ & 26.8810  & 4.02\%  & $\sim$ & 33.7746 & 8.84\%  & $\sim$ &27.7087  \\
\cmidrule(lr){2-12}
&   {POMO-random} & 23.2016 & 5.59\%  & $\sim$ & 28.1393 & 8.89\%  & $\sim$ & 35.6822 & 14.99\%  & $\sim$ &29.0077  \\
\cmidrule(lr){2-12}
&   AMDKD-POMO  & 22.7842 & 3.69\%  & $\sim$ & 27.4462 & 4.68\%  & $\sim$ & 34.0650 & 9.78\%  & $\sim$ &28.0985 \\
&   Omni-POMO$^\ddagger$  & 22.6562 & 3.11\%  & 13s & 26.8707 & 3.98\%  & 38s & 33.1435 & 6.81\%  & 4m & 27.5568 \\
\cmidrule(lr){2-12}
&   Ours-POMO-inter  & 22.4981 & 2.39\%  & 12s & \textbf{\emph{26.7699}} & \textbf{\emph{3.59\%}}  & 38s & 33.2138 & 7.04\%  & 3m & 27.4939 \\
&   Ours-POMO-intra & \textbf{22.4523} & \textbf{2.18\%}  & $\sim$ & \textbf{26.6468}& \textbf{3.12\%}  & $\sim$ &  33.0600 & 6.54\%  & $\sim$ &\textbf{\emph{27.3864}} \\
\cmidrule(lr){2-12}
\cmidrule(lr){2-12}
&   ELG-60 & 23.2704 & 5.90\%  & 14s & 27.9584 & 8.19\%  &39s & 35.4367 & 14.20\%  & 2.4m & 28.8885  \\
&   ELG-100 & 22.7460 & 3.52\%  & $\sim$ & 27.1748 &  5.16\%  & $\sim$ & 33.4630 & 7.84\%  & $\sim$ & 27.7946 \\
&   ELG-150 & 22.6514 & 3.08\%  & $\sim$ & 26.9530 & 4.30\%  &$\sim$ & 33.0651 & 6.56\% & $\sim$ & 27.5565 \\
\cmidrule(lr){2-12}
&    {ELG-random} & 22.7191 & 3.39\%  & $\sim$ & 27.1218 & 4.95\%  & $\sim$ & 33.4101 & 7.67\%  & $\sim$ & 27.7503  \\
\cmidrule(lr){2-12}
&   Ours-ELG-inter & 22.6303 & 2.99\%  & $\sim$ & 26.8971 &4.08\%  & $\sim$ & \textbf{\emph{32.8841}} & \textbf{\emph{5.97\%}}  & $\sim$ & 27.4705 \\
&   Ours-ELG-intra & 22.6052 & 2.87\%  & $\sim$ & 26.8063 & 3.73\%  & $\sim$ & \textbf{32.6863} & \textbf{5.45\%}  & $\sim$ & \textbf{27.3659}  \\
\bottomrule
\end{tabular}
\end{threeparttable}
}
\label{tab:unseen_sizes}
\end{table*}
\subsection{Comparison analysis}

We first verify the effectiveness of our approach on seen sizes during training for both TSP and CVRP, and the results are displayed in Table \ref{tab:main}. Specifically, we compare our approach with 1) highly specialized VRP solvers: Concorde~\citep{concorde} and LKH3~\citep{lkh3} for TSP, the hybrid genetic search (HGS)~\citep{vidal2022hybrid} and LKH3 for CVRP; 2) POMO-based methods, including the original POMO~\citep{kwon2020pomo}, AMDKD-POMO~\citep{bi2022learning} and Omni-POMO~\citep{zhou2023towards}; 3) recent learning-oriented routing solver ELG (specialized for enhancing generalization on complex node distributions and large problem sizes)~\citep{gao2023towards} for both TSP and CVRP. For POMO and ELG, we retrain the model on each problem size with equal epochs as our approach for a fair comparison, e.g., \emph{POMO-60} signifying the model trained on size 60, where POMO-random refers to the model trained on instances of random sizes within our training size range. AMDKD-POMO improved the cross-distribution generalization of POMO via knowledge distillation, where we retrain it following our training setups by tailoring teacher models to align with our exemplar sizes. Besides, we also show the results of its open-sourced pretrained models on the largest available sizes, i.e., AMDKD-POMO$^*$. Furthermore, Omni-POMO is a recent meta-learning framework to improve generalization across size and distribution of POMO, where we report their results by directly using their open-sourced pretrained models. Regarding our approach, two distinct variations with inter-task and intra-task regularization schemes are denoted as \emph{Ours-inter} and \emph{Ours-intra}, respectively. Every method is assessed using data augmentation of POMO. The total inference time is reported for all methods, i.e., GPU time for deep models and CPU time for traditional solvers.

From Table~\ref{tab:main}, we observe that our approach with intra-task regularization slightly outperforms the inter-task regularization variant on smaller problem sizes (e.g., 60 and 100 for TSP), but the other way round on larger sizes (e.g., 150 for TSP) for both Ours-POMO and Ours-ELG on TSP and CVRP. This is reasonable since intra-task regularization concentrate more on efficiently learning the latest larger sizes. Additionally, when compared to the original POMO and ELG models trained on a specific size, both Ours-POMO and Ours-ELG (incorporating either inter-task or intra-task regularization) exhibit competitive performance on those specific sizes for both TSP and CVRP. However, they significantly outperform the original POMO and ELG models in terms of average objective values across multiple problem sizes (see the final column). Regarding POMO-based methods, while specially designed to enhance the cross-distribution generalization of POMO, AMDKD-POMO$^*$ still suffers from the cross-size generalization issue. Furthermore, both Ours-POMO-inter and Ours-POMO-intra outperform POMO-random, AMDKD-POMO and Omni-POMO across all sizes for both TSP and CVRP with comparable inference time, even if Omni-POMO utilizes training instances with larger upper sizes (i.e., 200). Regarding ELG-based methods, while ELG models trained on specific sizes deliver slightly inferior performance to their POMO counterparts (e.g., POMO-60 vs. ELG-60), they demonstrate superior generalization performance on larger sizes (e.g., 150). This outcome is reasonable, as ELG employs an early stopping mechanism during POMO training to mitigate over-fitting, enhancing generalization at the potential expense of peak performance on those specific training sizes. Both Ours-ELG-inter and Ours-ELG-intra surpass ELG-random across all sizes for both TSP and CVRP, achieving better overall generalization (refer to the final column).

\subsection{Generalization analysis}

We further evaluate all methods on unseen larger sizes and gathered the results in Table~\ref{tab:unseen_sizes}. As revealed, the cross-size generalization issue of AMDKD-POMO$^*$ is more pronounced, which leads to a substantial deterioration in performance. Ours-POMO-inter surpasses POMO-random, POMO-60, POMO-100 and AMDKD-POMO across all sizes for both TSP and CVRP, and achieves competitive performance to POMO-150 (for CVRP200) and Omni-POMO (for TSP500 and CVRP500). Leveraging intra-task regularization to prioritize the learning of the latest larger sizes, Ours-POMO-intra further outperforms POMO-150 and Omni-POMO across all sizes for both TSP and CVRP. It is worth noting that Omni-POMO is trained on a broader range of sizes (including larger ones up to 200), which inherently offers Omni-POMO the potential for superior performance on larger sizes. Focusing on enhancing generalization on large problem sizes, ELG models trained on three specific sizes outperform their POMO counterparts (e.g., POMO-150 vs. ELG-150) on size 500 for both TSP and CVRP, and also exhibit significantly superior overall generalization performance in terms of average objective values across multiple problem sizes. Despite the superiority of ELG, both Ours-ELG-inter and Ours-ELG-intra still surpass ELG-60, ELG-100, ELG-150 and ELG-random across all sizes for both TSP and CVRP. Notably, Ours-ELG-intra consistently achieves lowest average objective values across the three sizes compared to all other neural baselines for both TSP and CVRP, which demonstrates the effectiveness of our approach in enhancing cross-size generalization of a backbone model.

\begin{table}
\caption{Generalization performance on instances (50\! $\leq\! N\!\leq$\! 500) from benchmark instances.}
\centering
\resizebox{0.75\textwidth}{!}{%
\begin{threeparttable}
\begin{tabular}{c|cccccc}
\toprule
& POMO-60 & POMO-100& POMO-150 & AMDKD-POMO & Omni-POMO & Ours-Inter  \\
\midrule
TSPLIB &  9.71\% & 4.49\%  &  4.18\% & 5.17\%  &    3.11\% &  4.07\% \\
CVRPLIB & 13.59\%  &  12.30\% & 9.21\%  &  7.09\%   &  5.83\% & 5.45\%   \\
\bottomrule
\end{tabular}
\end{threeparttable}
}
\label{tab:benchmark}
 \vspace{-10pt}
 \vspace{-0.1cm}
\end{table}


\begin{wraptable}{r}{0.55\textwidth}
    \vspace{-10pt}
\caption{{Ablation study on TSP.}}
\vspace{-5pt}
\centering
\resizebox{0.55\textwidth}{!}{
\begin{tabular}{ccc|cc|cc|cc}
\toprule
 & & & \multicolumn{2}{c|}{{N=60}} & \multicolumn{2}{c|}{{N=100}} & \multicolumn{2}{c}{{N=150}} \\
 ER & Inter-task & Intra-task  & Obj. & Gap & Obj. & Gap & Obj. & Gap  \\ 
\midrule
$\times$ & $\times$  & $\times$ & 6.1886  &  0.25\%  &  7.7898 &  0.32\%  & 9.3974 & 0.55\%   \\
$\times$ &  $\checkmark$ & $\times$ & {6.1805}  &  0.12\% & 7.7831 &  0.24\%   & 9.3938 & 0.51\%  \\
$\times$ & $\times$ & $\checkmark$ & 6.1809  &  0.13\% &  7.7829 &  0.24\%    & 9.3885 & 0.45\%   \\
$\checkmark$ &  $\times$  & $\times$ & 6.1789  &  0.10\%  &  7.7860 &  0.28\%& 9.3932 & 0.50\%   \\
$\checkmark$ & $\checkmark$ & $\times$ & 6.1758  &  0.05\% & 7.7775 &  0.17\%   & 9.3883 & 0.45\%  \\
$\checkmark$ & $\times$ & $\checkmark$ & 6.1758  &  0.05\% & 7.7764 &  0.15\%   & 9.3820 & 0.38\%  \\
\bottomrule
\end{tabular}}
\label{tab:ablation_tsp}
 \vspace{-10pt}
\end{wraptable}

We further extend the evaluation to realistic data taken from ``World TSP", which is available at~\citep{realdata} to show that both Ours-POMO and Ours-ELG consistently surpass their respective backbone baselines, i.e., POMO-150 and ELG-150, which further showcases the effectiveness of our approach. 

\subsection{Ablation study}
In Table \ref{tab:ablation_tsp}, we conduct an ablation study to clarify the effectiveness of each component of our approach on TSP, where only one regularization scheme can be used in our approach to keep a stable update of the exemplar model. The markers “$\checkmark$” and “$\times$” denote the utilization or exclusion of the corresponding component, respectively. The gaps are calculated based on the solutions acquired by Concorde in Table \ref{tab:main}. As exhibited, experience replay, inter-task and intra-task regularization schemes contribute to the reduction of objective values and optimality gaps across all sizes, affirming their effectiveness in enhancing cross-size generalizability. Further combining them together, both Ours-inter and Ours-intra (last two rows) achieve better performance. 



\section{Conclusions and future work}
\label{sec:conclusion}
This paper presents a continual learning based framework to foster the cross-size generalization of deep models for VRPs. We leverage either inter-task or intra-task regularization scheme to retain the valuable insights derived from previously trained exemplar models for facilitating subsequent training. To mitigate the catastrophic forgetting, we exploit the experience replay to revisit instances of formerly trained smaller sizes. Results show that our approach not only significantly strengthens the cross-size generalization performance, but also delivers predominantly superior performance to state-of-the-art deep models specialized for the generalizability enhancement. 


Scaling up to substantially large problem instances is important for future research. Bolstered by the superior cross-size generalization capacity, we will further improve the continual learning framework to train reliable deep models for handling large-scale VRPs, e.g., in a divide-and-conquer manner. Additionally, explicitly enhancing the cross-distribution generalization in the proposed CL
framework could further unleash the potential of our approach in real-world applications.

\bibliography{iclr2024_conference}
\bibliographystyle{iclr2024_conference}


\end{document}